\documentclass{wseas}
\usepackage{ragged2e}
\usepackage{booktabs}
\usepackage{multirow}
\usepackage{graphicx}
\usepackage{amsmath}
\usepackage{soul}
\usepackage{algorithm}
\usepackage{algpseudocode}

\author{
LUYAO WANG \\
University of Malaya\\
Malaysia
}

\title{Clustering-Enhanced Domain Adaptation for Cross-Domain Intrusion Detection in Industrial Control Systems}

\begin{document}
\twocolumn[
  \begin{@twocolumnfalse}
    \maketitle
    \begin{abstract}
Industrial control systems operate in dynamic environments where traffic distributions vary across scenarios, labeled samples are limited, and unknown attacks frequently emerge, posing significant challenges to cross-domain intrusion detection. To address this issue, this paper proposes a clustering-enhanced domain adaptation method for industrial control traffic. The framework contains two key components. First, a feature-based transfer learning module projects source and target domains into a shared latent subspace through spectral-transform-based feature alignment and iteratively reduces distribution discrepancies, enabling accurate cross-domain detection. Second, a clustering enhancement strategy combines K-Medoids clustering with PCA-based dimensionality reduction to improve cross-domain correlation estimation and reduce performance degradation caused by manual parameter tuning. Experimental results show that the proposed method significantly improves unknown attack detection. Compared with five baseline models, it increases detection accuracy by up to 49\%, achieves larger gains in F-score, and demonstrates stronger stability. Moreover, the clustering enhancement strategy further boosts detection accuracy by up to 26\% on representative tasks. These results suggest that the proposed method effectively alleviates data scarcity and domain shift, providing a practical solution for robust cross-domain intrusion detection in dynamic industrial environments.
    \end{abstract}


\begin{dates}
{\break
\color{red}
Received: May 31, 2019. Revised: May 4, 2020. Accepted: May 22, 2020. Published: May 29, 2020
\break(WSEAS will fill these dates in case of final acceptance, following strictly our data base and possible email
communication)}
\end{dates}
  \end{@twocolumnfalse}
\vspace{2ex} 
]

\section{Introduction}
Although deep learning has achieved remarkable success in pattern recognition~\cite{wang2024ssdag}, representation learning~\cite{mei2023udpreg,wang2023diga,ren2024pointcmae,wang2023zeroreg}, medical image analysis and healthcare intelligence~\cite{chang2025organagents,wang2025mbtpolyp,wang2024uhrmlp,ma2024llmmorph,xu2025scmllm}, as well as 3D understanding and retrieval tasks~\cite{nie2019hgan,li2025freeinsert,wang2024uvmap,gao2025pointpc,chang2024shapekg,gao2025pointpc,liang2020lstm3d,nie2019characteristic,10684147}, it has also shown strong potential in security-related applications and generative modeling~\cite{wang2023turn,wang2025fully,rigo2026pocidiff}. Nevertheless, its application to industrial control system intrusion detection remains limited in practice, mainly due to scarce labeled data, heterogeneous traffic distributions, and the dynamic nature of industrial environments~\cite{umer2022mlics,gauthama2021icsml,kheddar2023dtl}.

Industrial Control Systems (ICSs) have become indispensable to modern critical infrastructure, supporting a wide range of industrial sectors such as petrochemical production, water treatment, natural gas transportation, and electric power systems~\cite{umer2022mlics,gauthama2021icsml}. By integrating industrial process control with communication networks and intelligent management technologies, ICSs enable real-time monitoring, automated regulation, and efficient coordination of large-scale industrial operations. As industrial environments continue to evolve toward higher levels of automation, connectivity, and intelligence, their dependence on ICSs has become deeper than ever. Although this transformation greatly improves productivity, flexibility, and management efficiency, it also introduces new cybersecurity risks that directly threaten the reliability and safety of industrial operations~\cite{umer2022mlics,kravchik2018icscnn}.

\begin{figure}[t]
    \centering
\textbf{}    \includegraphics[width=\linewidth]{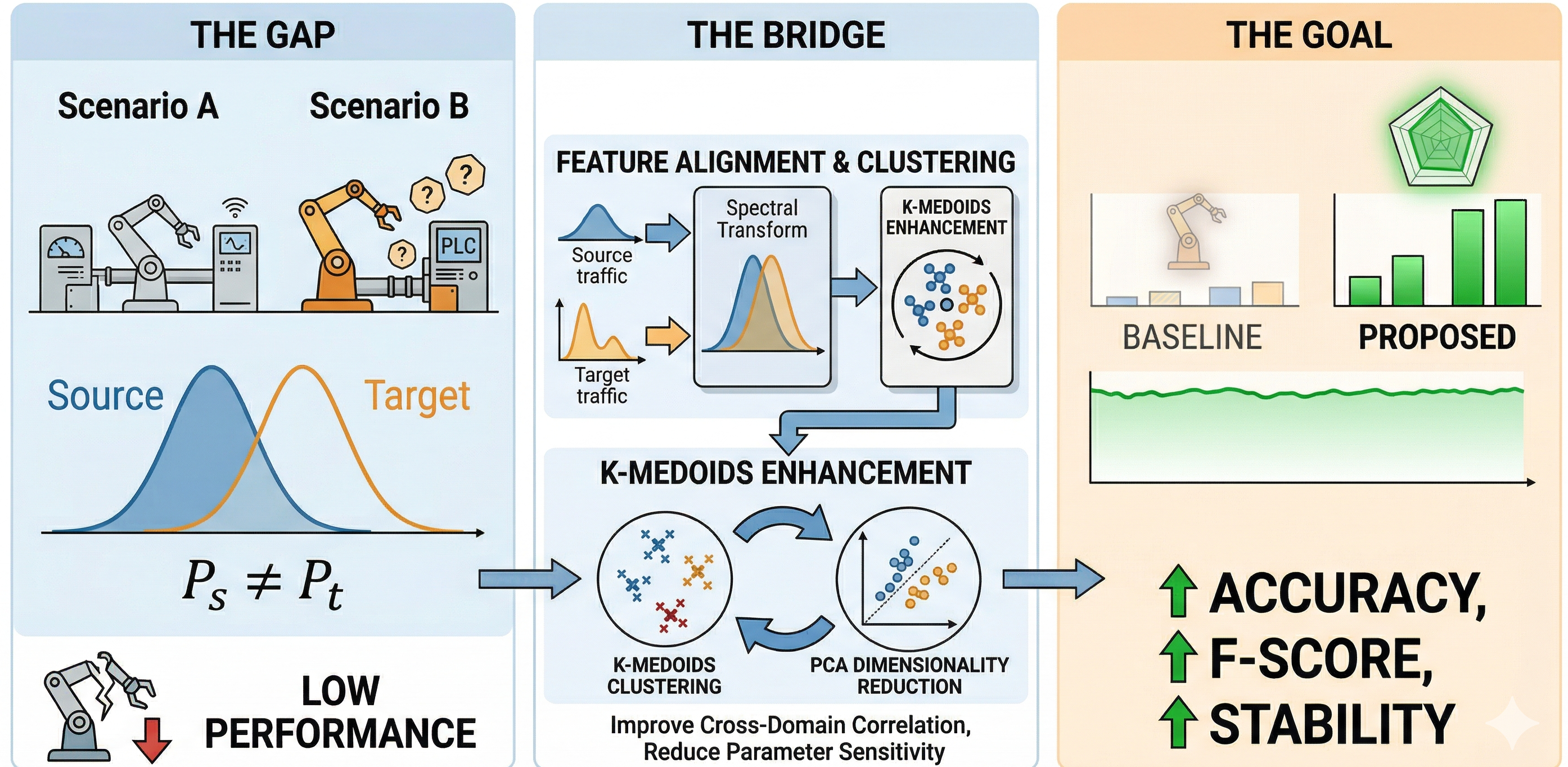}
    \caption{
Teaser of the proposed clustering-enhanced domain adaptation framework for industrial control intrusion detection.}
    \label{fig:teaser}
\end{figure}

Compared with conventional IT systems, industrial control networks exhibit several unique characteristics that make intrusion detection substantially more challenging~\cite{umer2022mlics,gauthama2021icsml}. First, the distributed and intelligent evolution of industrial systems has accelerated the adoption of remote access and interconnected communication mechanisms, thereby exposing vulnerabilities in industrial devices, protocols, and software to potential attackers. Second, ICSs operate under strict real-time communication requirements. In practical settings, traffic from distributed field devices may rapidly converge to central control nodes during specific operational periods, resulting in high-speed, bursty, and highly concurrent communication patterns. These properties can be statistically similar to malicious traffic, allowing attack behaviors to be concealed within normal operational communication and making accurate detection difficult~\cite{gauthama2021icsml,kravchik2018icscnn}. In addition, industrial communication networks typically involve heterogeneous devices, complex monitoring equipment, large-scale topologies, and rich yet highly structured traffic features. As a result, many detection strategies designed for conventional Internet or IoT environments cannot be directly transferred to industrial scenarios~\cite{umer2022mlics,kheddar2023dtl}.

To protect industrial systems from cyber threats, a variety of intrusion detection approaches have been studied, including host-based, protocol-based, and traffic-based methods~\cite{umer2022mlics,gauthama2021icsml}. Among them, traffic-based intrusion detection has become one of the most mainstream paradigms and is the focus of this work. The reason is that industrial traffic contains abundant behavioral information, such as communication sessions, packet statistics, and temporal variation patterns. By extracting and analyzing these features, an intrusion detection model can learn discriminative patterns that distinguish normal communication from malicious behavior. In recent years, machine learning and deep learning techniques have significantly improved feature mining and attack classification, enabling more powerful intrusion detection systems for industrial environments~\cite{kravchik2018icscnn,wang2023diga,wang2025fully}. However, most of these methods rely on two assumptions: the availability of sufficient high-quality labeled training data and the consistency of feature distributions between training and testing samples. In real industrial deployments, both assumptions are often violated~\cite{umer2022mlics,kheddar2023dtl,gauthama2021icsml}.

A fundamental challenge is that industrial communication environments are inherently dynamic. Changes in monitored devices, deployment locations, manufacturers, control processes, communication patterns, and threat types can all cause significant shifts in data distribution~\cite{umer2022mlics,kheddar2023dtl}. Consequently, a detector trained on one industrial system may perform well in its original environment but degrade substantially when applied to another system or even to a later stage of the same system after structural changes occur. This problem becomes particularly severe in unknown attack detection. For example, a model trained on historical traffic from one scenario may fail to recognize new attack behaviors introduced by newly connected equipment or newly exposed vulnerabilities. Similarly, when a new industrial system is deployed, there is often insufficient labeled traffic to train a dedicated intrusion detection model from scratch, forcing practitioners to reuse models trained on other systems. Because industrial traffic distributions are not consistent across systems, such reused models often exhibit weak generalization ability in the new environment. These practical scenarios highlight the importance of cross-domain intrusion detection, namely, enabling a model trained on a labeled source domain to generalize effectively to an unlabeled or sparsely labeled target domain~\cite{zhao2019unknownattack,ganin2016dann,kheddar2023dtl}.

Unfortunately, collecting large-scale, high-quality, and continuously updated labeled industrial traffic data is extremely difficult. Industrial control traffic is scarce and valuable, and relevant data are often distributed across different network layers and operational stages. Moreover, traffic acquisition, cleaning, screening, annotation, model retraining, and redeployment all incur substantial time and economic costs. As a result, conventional supervised retraining is often impractical in real deployments~\cite{umer2022mlics,gauthama2021icsml}. Therefore, reducing dependence on labeled target-domain traffic while building adaptable and accurate intrusion detection models for dynamic industrial environments has become an urgent research problem in industrial cybersecurity~\cite{kheddar2023dtl,zhao2019unknownattack}.

Transfer learning and domain adaptation provide a promising solution to this challenge~\cite{ganin2016dann,zhao2019unknownattack,kheddar2023dtl}. Instead of training a detector exclusively within a single fixed environment, transfer learning aims to reuse knowledge learned from a labeled source domain and transfer it to a related but different target domain. This paradigm is particularly attractive for ICS intrusion detection, where labeled attack samples in the target domain are scarce or unavailable, while certain structural similarities may still exist across industrial systems~\cite{kheddar2023dtl,zhao2019unknownattack}. Existing studies have shown that transfer learning can alleviate the shortage of labeled datasets and improve cross-domain detection performance~\cite{ganin2016dann,zhao2019unknownattack}. Nevertheless, directly applying transfer learning to industrial traffic remains nontrivial. Source and target domains may differ in feature dimensionality, statistical properties, and noise levels. Moreover, industrial traffic often contains outliers and complex attack patterns, which hinder reliable correspondence estimation between domains~\cite{gauthama2021icsml,kheddar2023dtl}. Therefore, an effective adaptation framework for ICS intrusion detection must address not only feature distribution mismatch, but also stable and accurate source--target correlation modeling under heterogeneous and noisy conditions.

To address these challenges, this paper proposes a clustering-enhanced domain adaptation framework for industrial control traffic intrusion detection. The proposed method consists of two tightly coupled components. The first is a feature-based transfer learning module that maps the source and target domains into a shared latent subspace through spectral-transform-based feature alignment. By iteratively reducing the discrepancy between corresponding instances across domains, this module learns a representation that is more suitable for cross-domain intrusion detection. The second is a clustering enhancement strategy that improves the reliability of source--target matching before and during adaptation. Specifically, K-Medoids clustering is employed to partition source and target samples~\cite{kaufman1990pam}, while PCA is used to homogenize heterogeneous feature spaces so that meaningful distance-based correlation analysis can be performed~\cite{pearson1901pca}. Compared with direct feature transfer alone, the proposed clustering-enhanced mechanism improves the reliability of domain correlation estimation and reduces performance degradation caused by manual parameter tuning, thereby leading to more stable adaptation behavior.

The effectiveness of the proposed framework is validated on industrial traffic datasets collected from natural gas pipeline systems and water storage control systems. Experimental results show that the proposed method substantially improves unknown attack detection performance in cross-domain scenarios. Compared with five common baseline models, it improves detection accuracy by up to 49\%, achieves even larger gains in F-score, and exhibits stronger stability. In addition, the clustering enhancement strategy further improves detection accuracy by up to 26\% on representative tasks. These findings demonstrate that incorporating clustering structure into feature-based domain adaptation is a practical and effective way to mitigate data scarcity and domain shift in industrial intrusion detection.

The main contributions of this paper are summarized as follows:
\begin{itemize}
    \item We investigate the problem of cross-domain intrusion detection in industrial control systems and analyze the joint impact of data scarcity, dynamic environments, and domain shift on the performance of conventional intrusion detection models.
    \item We propose a clustering-enhanced domain adaptation framework that integrates feature-based transfer learning with K-Medoids-based clustering enhancement and PCA-based feature homogenization for industrial traffic data.
    \item We demonstrate through experiments on industrial traffic datasets that the proposed method consistently improves detection accuracy, F-score, and stability, showing its effectiveness for unknown attack detection in dynamic industrial environments.
\end{itemize}
\section{Research Methodology}

\begin{figure*}[t]
    \centering
    \includegraphics[width=0.8\textwidth]{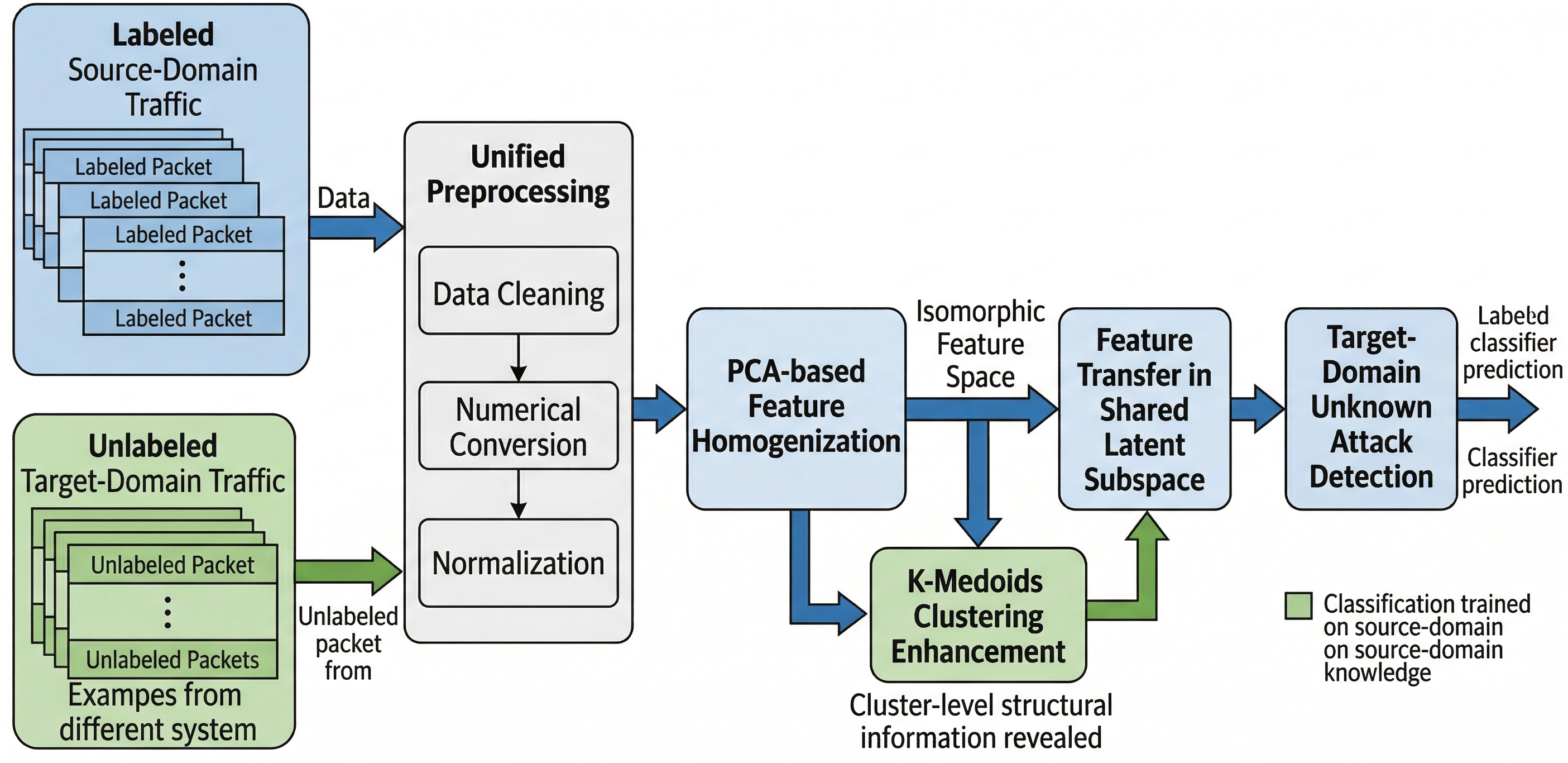}
    \caption{
    Overview of the proposed clustering-enhanced domain adaptation framework for cross-domain intrusion detection in industrial control systems.
    The left part illustrates the challenge of domain shift and unknown attacks in dynamic industrial environments.
    The middle part presents the proposed method, which combines PCA-based feature homogenization, K-Medoids-guided structural correspondence estimation, and spectral-transform-based feature alignment.
    The right part summarizes the effectiveness of the proposed framework in improving unknown attack detection accuracy, F-score, and stability over baseline methods.
    }
    \label{fig:framework}
\end{figure*}

This section presents the proposed clustering-enhanced domain adaptation framework for cross-domain intrusion detection in industrial control systems~\cite{kheddar2023dtl,gauthama2021icsml}. The objective is to transfer discriminative knowledge from a labeled source domain to an unlabeled target domain under distribution mismatch, so that the detector can maintain reliable performance in dynamic industrial environments~\cite{ganin2016dann,zhao2019unknownattack,kheddar2023dtl}. In this study, intrusion detection is formulated as a binary classification problem, where each traffic sample is categorized as either normal or malicious. The proposed framework consists of two tightly coupled components: a feature-based transfer learning module for latent-space alignment and a clustering enhancement module for structure-aware source--target correspondence estimation. As illustrated in Fig.~\ref{fig:framework}, the overall pipeline starts from heterogeneous source-domain and target-domain traffic, then performs feature-space homogenization and clustering-guided correspondence estimation, and finally learns a transferable latent representation for target-domain intrusion detection.

\subsection{Problem Formulation}

Let the labeled source domain be defined as
\[
\mathcal{D}_s=\{(x_i^s,y_i^s)\}_{i=1}^{n_s},
\]
where \(x_i^s \in \mathbb{R}^{d_s}\) denotes the \(i\)-th source-domain sample and \(y_i^s \in \{0,1\}\) is its label, with \(0\) representing normal traffic and \(1\) representing malicious traffic. Let the unlabeled target domain be defined as
\[
\mathcal{D}_t=\{x_j^t\}_{j=1}^{n_t},
\]
where \(x_j^t \in \mathbb{R}^{d_t}\) denotes the \(j\)-th target-domain sample. In practical industrial scenarios, \(\mathcal{D}_s\) and \(\mathcal{D}_t\) are collected from different control environments, such as natural gas pipeline systems and water storage control systems. As a result, the two domains may differ in both feature dimensionality and statistical distribution~\cite{bendavid2010domain,gauthama2021icsml,kheddar2023dtl,chen2023crossdomain}, i.e.,
\[
d_s \neq d_t,\qquad P_s(X,Y)\neq P_t(X,Y).
\]

The goal is to learn a mapping \(f(\cdot)\) and a classifier \(g(\cdot)\) using labeled source-domain data and unlabeled target-domain data such that the target-domain labels can be predicted accurately:
\[
\hat{y}_j^t = g(f(x_j^t)), \qquad j=1,\dots,n_t.
\]
To achieve this, the proposed framework aims to reduce cross-domain discrepancy while preserving the discriminative structure required for industrial intrusion detection. From the perspective of Fig.~\ref{fig:framework}, this objective is achieved by combining global latent-space transfer with local structure-aware matching between the two domains.

\subsection{Data Preprocessing and Feature Space Homogenization}

Before domain adaptation, the source-domain and target-domain traffic data are preprocessed in a unified manner. The preprocessing pipeline includes erroneous-label correction, numerical encoding of categorical attributes, and feature standardization. For each domain, feature standardization is performed as
\[
\tilde{x} = \frac{x-\mu}{\sigma},
\]
where \(\mu\) and \(\sigma\) denote the feature-wise mean and standard deviation, respectively. This step ensures that variables with different physical scales can be represented consistently.

A key challenge in industrial traffic adaptation is that the source and target domains may not share the same feature dimensionality. Direct similarity computation and cluster analysis are therefore unreliable under heterogeneous feature spaces. To address this issue, principal component analysis (PCA) is used to project both domains into a common \(d\)-dimensional isomorphic space~\cite{pearson1901pca,jolliffe2002pca}:
\[
z_i^s = W_s^\top \tilde{x}_i^s,\qquad z_j^t = W_t^\top \tilde{x}_j^t,
\]
where \(W_s \in \mathbb{R}^{d_s \times d}\) and \(W_t \in \mathbb{R}^{d_t \times d}\) are PCA projection matrices for the source and target domains, respectively, and \(d \ll \min(d_s,d_t)\). After transformation, both domains are represented as
\[
\mathcal{Z}_s=\{(z_i^s,y_i^s)\}_{i=1}^{n_s},\qquad
\mathcal{Z}_t=\{z_j^t\}_{j=1}^{n_t},
\]
which provides a unified representation basis for clustering and feature transfer. This preprocessing and homogenization stage corresponds to the first key step shown in Fig.~\ref{fig:framework}, where heterogeneous industrial traffic is converted into a comparable representation space for subsequent adaptation.

\subsection{Feature-Based Transfer Learning}

Feature-based transfer learning forms the core of the proposed framework~\cite{pan2009tca,ganin2016dann,zhao2019unknownattack,kheddar2023dtl}. Its purpose is to learn a shared latent representation in which the discrepancy between source and target domains is reduced, while the class structure relevant to intrusion detection is preserved.

Let
\[
h_i^s = A^\top z_i^s,\qquad h_j^t = A^\top z_j^t,
\]
where \(A \in \mathbb{R}^{d \times m}\) is a latent projection matrix and \(m\) is the dimensionality of the latent subspace. The transformed feature sets are denoted by
\[
\mathcal{H}_s=\{(h_i^s,y_i^s)\}_{i=1}^{n_s},\qquad
\mathcal{H}_t=\{h_j^t\}_{j=1}^{n_t}.
\]

To make the learned representation transferable, the latent mapping is optimized by simultaneously preserving source-domain discriminability and reducing source--target discrepancy~\cite{bendavid2010domain,pan2009tca,ganin2016dann,zhao2019unknownattack}. A generic objective can be written as
\[
\min_{A}\; \mathcal{L}_{cls}(\mathcal{H}_s,Y_s)
+ \lambda \mathcal{L}_{da}(\mathcal{H}_s,\mathcal{H}_t)
+ \gamma \|A\|_F^2,
\]
where \(\mathcal{L}_{cls}\) is the source-domain classification loss, \(\mathcal{L}_{da}\) is the domain adaptation loss measuring the discrepancy between the two domains in the latent space, and \(\lambda,\gamma>0\) are trade-off parameters.

In this work, the alignment process is implemented in a spectral-transform-based manner. The latent projection iteratively reduces the distance between structurally corresponding source and target samples:
\[
\mathcal{L}_{da}=\sum_{(i,j)\in \Omega}\|h_i^s-h_j^t\|_2^2,
\]
where \(\Omega\) denotes the set of cross-domain correspondences estimated by the clustering enhancement module. By minimizing this term, the latent representation gradually captures the domain-invariant structure shared by different industrial systems. A classifier \(g(\cdot)\) is then trained on the adapted source-domain representation and used to infer target-domain labels. As summarized in Fig.~\ref{fig:framework}, this module serves as the feature transfer backbone of the proposed method, while its optimization is guided by the structural correspondence discovered by the clustering enhancement module.

\subsection{Clustering Enhancement Strategy}

\begin{figure}[t]
    \centering
    \includegraphics[width=\linewidth]{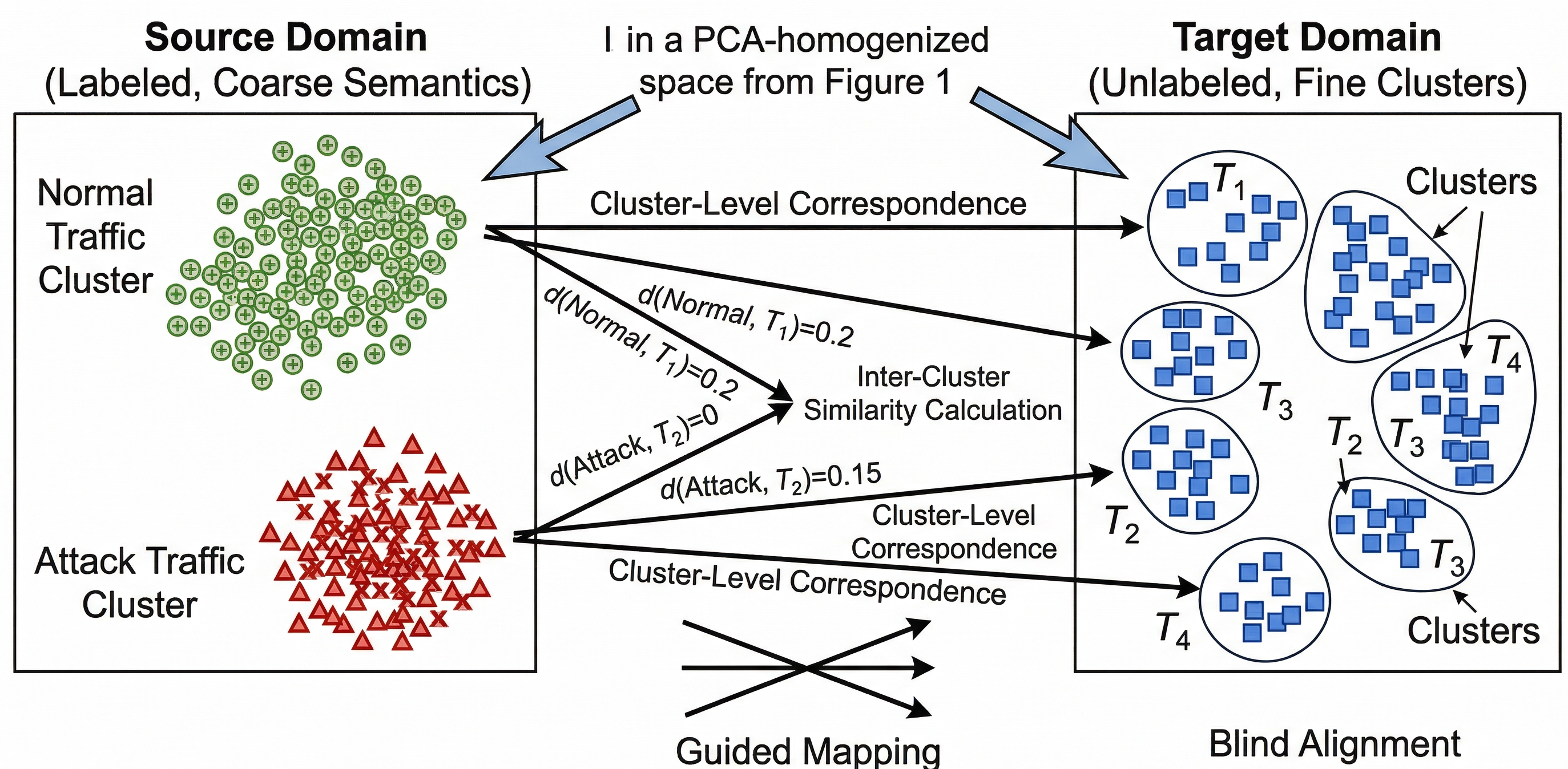}
    \caption{
    Illustration of the clustering enhancement module. After PCA-based feature homogenization, source-domain and target-domain samples are partitioned into clusters by K-Medoids. Cluster-level similarity is then computed to establish source--target structural correspondence, which guides the subsequent latent-space alignment.
    }
    \label{fig:module}
\end{figure}

Although latent-space transfer learning reduces domain discrepancy, direct alignment of all samples may still be unreliable when industrial traffic contains noise, outliers, and heterogeneous attack patterns~\cite{gauthama2021icsml,kheddar2023dtl}. To improve source--target matching quality, a clustering enhancement strategy is introduced before latent-space alignment. The internal workflow of this module is illustrated in Fig.~\ref{fig:module}, where source-domain and target-domain samples are first organized into clusters and then matched through cluster-level similarity estimation.

\begin{algorithm}[t]
\caption{Clustering-Enhanced Domain Adaptation for ICS Intrusion Detection}
\label{method:code}
\textbf{Input:} Source domain $\mathcal{D}_s$, target domain $\mathcal{D}_t$

\textbf{Output:} Predicted target labels $\hat{Y}_t$

1. Preprocess source and target data

2. Apply PCA to obtain $\mathcal{Z}_s$ and $\mathcal{Z}_t$

3. Perform K-Medoids clustering on both domains

4. Compute cluster-level similarity and build correspondence set $\Omega$

5. Learn latent projection $A$ by minimizing the adaptation objective

6. Train classifier $g$ on adapted source features

7. Predict target labels $\hat{Y}_t = g(A^\top Z_t)$
\end{algorithm}

The complete computational procedure of the proposed method is summarized in Alg.~\ref{method:code}. It shows how preprocessing, PCA projection, K-Medoids clustering, structural correspondence construction, latent-space learning, and final prediction are integrated into a unified cross-domain intrusion detection pipeline.

After PCA transformation, the source and target samples are partitioned into \(K_s\) and \(K_t\) clusters, respectively:
\[
\mathcal{C}_s=\{C_1^s,C_2^s,\dots,C_{K_s}^s\},\qquad
\mathcal{C}_t=\{C_1^t,C_2^t,\dots,C_{K_t}^t\}.
\]
Each cluster is represented by its medoid~\cite{kaufman1990pam}:
\[
m_p^s \in C_p^s,\qquad m_q^t \in C_q^t,
\]
where \(m_p^s\) and \(m_q^t\) are actual samples minimizing the total within-cluster distance.

Cluster-level similarity between the \(p\)-th source cluster and the \(q\)-th target cluster is measured by
\[
S_{pq} = \exp\!\left(-\frac{\|m_p^s-m_q^t\|_2^2}{\tau}\right),
\]
where \(\tau>0\) is a scaling parameter. For each target-domain cluster \(C_q^t\), the most relevant source-domain cluster is selected as
\[
\pi(q)=\arg\max_{p} S_{pq}.
\]
This yields a coarse structural correspondence between the two domains:
\[
C_q^t \longleftrightarrow C_{\pi(q)}^s.
\]

Based on this cluster-level matching, sample-level correspondence candidates are further constructed within matched cluster pairs, and the resulting correspondence set \(\Omega\) is used to guide latent-space alignment. Compared with global sample-wise transfer, this strategy reduces unreliable matching caused by severe domain discrepancy and provides a more stable structural prior for adaptation. This mechanism is visually summarized in Fig.~\ref{fig:module} and operationally reflected in Steps 3--5 of Alg.~\ref{method:code}.

\subsection{K-Medoids for Noise-Robust Cluster Alignment}

The selection of clustering algorithm is particularly important for industrial traffic analysis because ICS datasets often contain noisy observations, rare attacks, and imbalanced traffic patterns~\cite{gauthama2021icsml,kheddar2023dtl}. Under such conditions, cluster quality can be severely affected if the clustering method is sensitive to outliers.

For this reason, K-Medoids is adopted instead of K-Means~\cite{kaufman1990pam,mushtaq2018pam}. Given a set of PCA-transformed samples \(\{z_i\}_{i=1}^n\), K-Medoids aims to minimize the total within-cluster dissimilarity:
\[
\min_{\{C_k\},\{m_k\}} \sum_{k=1}^{K}\sum_{z_i\in C_k}\|z_i-m_k\|_2,
\]
where \(m_k \in C_k\) is the medoid of cluster \(C_k\). Unlike K-Means, which uses the arithmetic mean as the cluster center, K-Medoids selects actual samples as representatives. This property makes it more robust to noise and outliers~\cite{kaufman1990pam,mushtaq2018pam}, and therefore more suitable for industrial traffic data with irregular attack behavior.

By applying K-Medoids in the PCA-homogenized feature space, the proposed framework obtains more reliable cluster partitions and more stable source--target structural correspondence, which in turn improves the effectiveness of subsequent domain adaptation. As reflected in Fig.~\ref{fig:module} and Alg.~\ref{method:code}, the clustering module plays a central role in bridging local structure discovery and global latent-space alignment.
\section{Experiments}

This section evaluates the proposed clustering-enhanced domain adaptation framework on cross-domain intrusion detection tasks in industrial control systems. We first describe the experimental setup, including baseline models, task construction, and evaluation metrics. We then report the results on unknown attack detection and further analyze the contribution of the clustering enhancement strategy.

\subsection{Experimental Setup}

In this study, industrial control intrusion detection is formulated as a binary classification problem, where each traffic sample is categorized as either normal or malicious. To evaluate the effectiveness of the proposed framework, five widely used machine learning models are adopted as baseline methods, including Random Forest (RF), Support Vector Machine (SVM), Naive Bayes Model (NBM), K-Nearest Neighbors (KNN), and Artificial Neural Network (ANN). Among them, ANN is implemented as a three-layer fully connected neural network using PyTorch, while the remaining models are implemented with the scikit-learn library.

The proposed clustering-enhanced domain adaptation framework is also implemented in PyTorch. To ensure a fair comparison, all methods are evaluated under the same source--target task settings and binary intrusion detection protocol. In preliminary experiments, a single decision tree consistently showed weaker performance than random forest under cross-domain industrial intrusion detection settings. Therefore, it is not included in the final comparison. This design allows the evaluation to focus on representative classical and neural baselines while avoiding redundant comparisons with clearly inferior variants.

To evaluate the performance of the proposed method on unknown attack detection, four cross-domain tasks are constructed based on two industrial control systems: a natural gas control system (G) and a water storage tank control system (W). The tasks are defined as:
\begin{itemize}
    \item DoS(G)$\rightarrow$NMRI(W),
    \item SMRI(G)$\rightarrow$MPCI(W),
    \item DoS(W)$\rightarrow$NMRI(G),
    \item SMRI(W)$\rightarrow$MPCI(G).
\end{itemize}
For example, in DoS(G)$\rightarrow$NMRI(W), the source domain consists of normal traffic and DoS attack traffic collected from the gas system, while the target domain consists of normal traffic and NMRI attack traffic collected from the water system. The remaining tasks are constructed in the same manner by exchanging the source and target systems as well as attack categories. These settings are designed to evaluate the ability of each method to generalize across both industrial environments and unseen attack patterns.

Such a protocol is particularly challenging because the source and target domains differ not only in attack category, but also in traffic semantics, payload characteristics, and control-process dynamics. Therefore, a detector that performs well under these settings must possess both feature transferability and robustness to unseen domain-specific variations. This makes the experimental design a meaningful testbed for evaluating cross-domain industrial intrusion detection capability rather than simple within-domain classification accuracy.

In addition to the main comparison, an ablation-style analysis is conducted to assess the contribution of the clustering enhancement strategy. Specifically, the proposed framework is compared with its counterpart without clustering enhancement on two representative tasks with clear domain discrepancy, namely DoS(G)$\rightarrow$NMRI(W) and DoS(W)$\rightarrow$NMRI(G). These tasks are selected because they reflect bidirectional transfer across distinct industrial systems and therefore provide a useful setting for analyzing the effect of structural alignment.

\subsection{Evaluation Metrics}

To evaluate detection performance, Accuracy (ACC) and F-score (F1) are adopted as the primary metrics. Accuracy measures the proportion of correctly classified samples and is defined as
\begin{equation}
\mathrm{ACC}=\frac{TP+TN}{TP+TN+FP+FN},
\end{equation}
where $TP$, $TN$, $FP$, and $FN$ denote true positives, true negatives, false positives, and false negatives, respectively.

F-score is used to provide a more balanced evaluation by jointly considering precision and recall:
\begin{equation}
\mathrm{F1}=\frac{2 \times \mathrm{Precision} \times \mathrm{Recall}}{\mathrm{Precision}+\mathrm{Recall}},
\end{equation}
where
\begin{equation}
\mathrm{Precision}=\frac{TP}{TP+FP}, \qquad
\mathrm{Recall}=\frac{TP}{TP+FN}.
\end{equation}

Since industrial intrusion detection may involve class imbalance and unknown attacks, F1 is particularly important for assessing the practical effectiveness of a detector. A model with high accuracy but low F1 may still be unreliable in practice, because it may fail to identify malicious traffic consistently or may produce excessive false alarms. In addition, ROC curves and the area under the ROC curve (AUC) are used for supplementary analysis of discriminative ability, where the horizontal axis corresponds to the false positive rate (FPR) and the vertical axis denotes the true positive rate (TPR). These complementary metrics provide a more comprehensive view of classifier behavior under different threshold settings.

\subsection{Results on Unknown Attack Detection}

Table~\ref{tab:acc_unknown_attack} and Table~\ref{tab:f1_unknown_attack} report the accuracy and F-score results of the proposed method and five baseline models on four cross-domain unknown attack detection tasks.

\begin{table}[t]
\centering
\caption{Accuracy of unknown attack detection on four cross-domain tasks.}
\label{tab:acc_unknown_attack}
\resizebox{\columnwidth}{!}{%
\begin{tabular}{lcccccc}
\toprule
Detection Task & KMM & RF & SVM & NBM & KNN & ANN \\
\midrule
DoS(G)$\rightarrow$NMRI(W)   & 0.85 & 0.47 & 0.51 & 0.34 & 0.47 & 0.52 \\
SMRI(G)$\rightarrow$MPCI(W)  & 0.87 & 0.49 & 0.52 & 0.37 & 0.48 & 0.55 \\
DoS(W)$\rightarrow$NMRI(G)   & 0.84 & 0.45 & 0.49 & 0.29 & 0.41 & 0.50 \\
SMRI(W)$\rightarrow$MPCI(G)  & 0.81 & 0.47 & 0.49 & 0.31 & 0.46 & 0.50 \\
\bottomrule
\end{tabular}%
}
\end{table}

\begin{table}[t]
\centering
\caption{F-score of unknown attack detection on four cross-domain tasks.}
\label{tab:f1_unknown_attack}
\resizebox{\columnwidth}{!}{%
\begin{tabular}{lcccccc}
\toprule
Detection Task & KMM & RF & SVM & NBM & KNN & ANN \\
\midrule
DoS(G)$\rightarrow$NMRI(W)   & 0.86 & 0.22 & 0.34 & 0.04 & 0.25 & 0.41 \\
SMRI(G)$\rightarrow$MPCI(W)  & 0.84 & 0.26 & 0.42 & 0.34 & 0.23 & 0.51 \\
DoS(W)$\rightarrow$NMRI(G)   & 0.85 & 0.42 & 0.27 & 0.02 & 0.46 & 0.26 \\
SMRI(W)$\rightarrow$MPCI(G)  & 0.80 & 0.24 & 0.09 & 0.15 & 0.48 & 0.42 \\
\bottomrule
\end{tabular}%
}
\end{table}

As shown in Table~\ref{tab:acc_unknown_attack}, the proposed framework consistently achieves the highest accuracy on all four cross-domain tasks. Its accuracy reaches 0.85, 0.87, 0.84, and 0.81, respectively, whereas the baseline methods remain substantially lower. The largest accuracy gain reaches 49\% under the selected settings, indicating that the proposed framework can effectively alleviate the performance degradation caused by domain shift and unknown attacks.

The superiority of the proposed method is even more evident in terms of F-score, as shown in Table~\ref{tab:f1_unknown_attack}. The proposed framework obtains F1 values of 0.86, 0.84, 0.85, and 0.80 on the four tasks, respectively, while the baseline models generally achieve much lower scores. This result suggests that the proposed method not only improves overall classification correctness, but also achieves a better balance between precision and recall. Such a property is especially important in industrial intrusion detection, where both false alarms and missed detections can lead to substantial operational risk.

\begin{figure}[!htp]
    \centering
    \includegraphics[width=\linewidth]{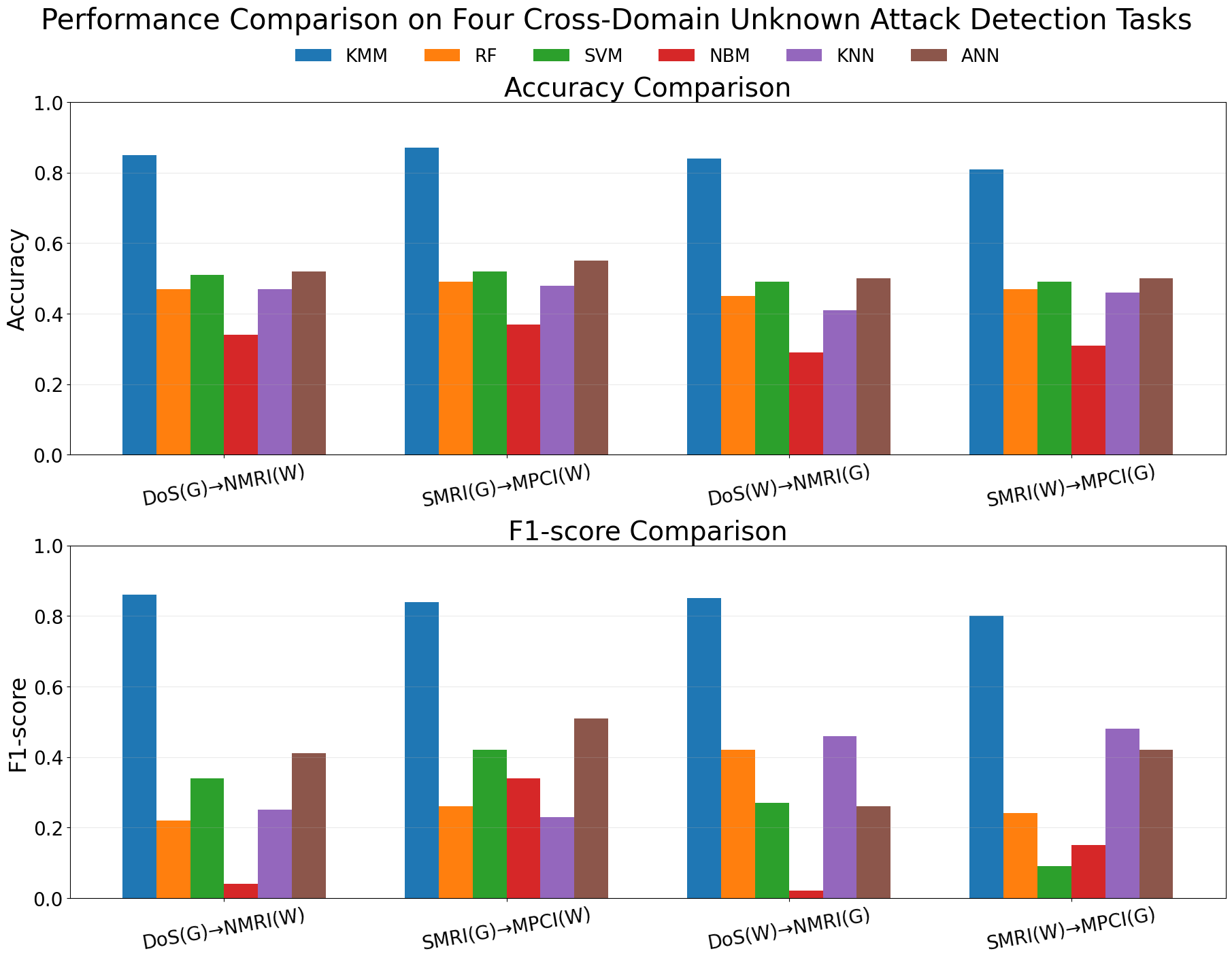}
    \caption{
    Visual comparison of the proposed framework and baseline models on four cross-domain unknown attack detection tasks. The upper panel reports the accuracy results, while the lower panel presents the F-score results. The proposed method consistently achieves the best performance across different transfer settings, demonstrating its superiority in both classification correctness and class-balanced detection effectiveness.
    }
    \label{fig:main_exp}
    \vspace{-3mm}
\end{figure}

Fig.~\ref{fig:main_exp} provides a more intuitive visualization of the comparative results reported in Table~\ref{tab:acc_unknown_attack} and Table~\ref{tab:f1_unknown_attack}. The upper panel shows the accuracy comparison, while the lower panel illustrates the F-score comparison across the four cross-domain tasks. It can be observed that the proposed KMM-based framework consistently outperforms all baseline methods under all transfer settings, and its advantage is particularly clear in terms of F-score. This further confirms that the proposed method is more effective in maintaining a balanced trade-off between precision and recall when detecting unknown attacks in heterogeneous industrial environments.

Another important observation from Fig.~\ref{fig:main_exp} is that the relative performance gap between the proposed framework and the baseline models remains stable across different source--target transfer directions. This suggests that the benefit of the proposed method does not rely on a single favorable task, but instead reflects a generally improved capability for cross-domain industrial intrusion detection. Moreover, the proposed framework shows stronger robustness on the more challenging reverse-transfer settings, which may be attributed to weaker source-domain informativeness and larger domain discrepancy.

These improvements can be attributed to two aspects of the proposed framework. First, feature-based transfer learning identifies a latent representation in which structurally related samples from different industrial domains become more comparable, thereby improving cross-domain generalization. Second, the clustering enhancement strategy provides a coarse structural prior for source--target matching, which reduces unreliable alignment and improves the robustness of feature transfer. Together, these two components enable the detector to exploit transferable knowledge while preserving task-relevant discrimination.

\begin{figure}[!htp]
    \centering
    \includegraphics[width=\linewidth]{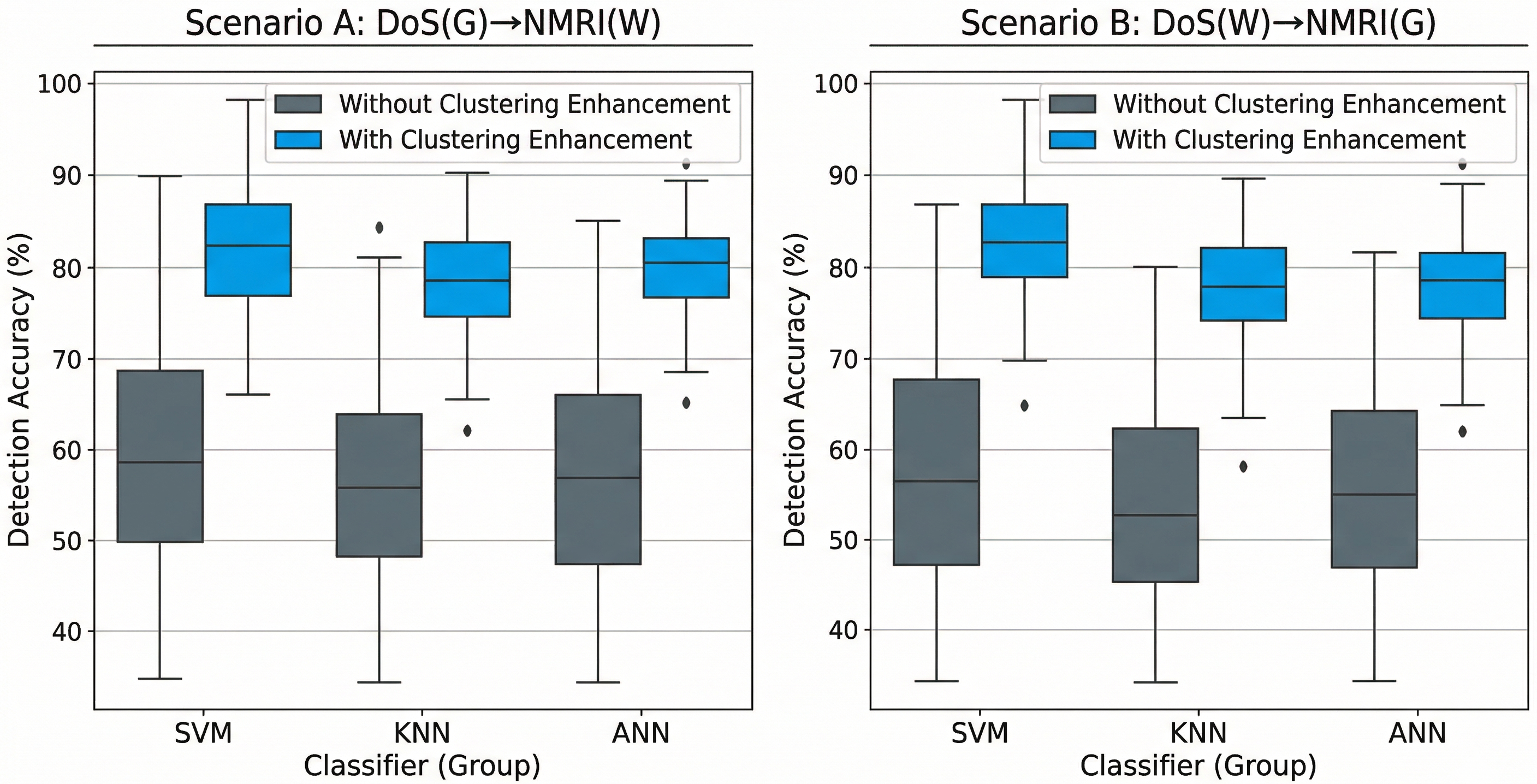}
    \caption{
    Performance comparison of the proposed method and baseline models on cross-domain unknown attack detection tasks. The results show that the proposed framework consistently achieves superior accuracy and F-score under different source--target transfer settings.
    }
    \label{fig:exp}
\end{figure}

Another noteworthy observation from Table~\ref{tab:acc_unknown_attack}, Table~\ref{tab:f1_unknown_attack}, and Fig.~\ref{fig:exp} is that the baseline models tend to perform better when G is used as the source domain than when W is used as the source domain. A plausible explanation is that the gas-system traffic contains richer payload and control-related information, such as pipeline pressure, PID gain parameters, and controller period settings, which provide more informative features for transferable representation learning. In contrast, the water-system traffic mainly reflects water-level setpoint information and may therefore contain less discriminative variation for cross-domain transfer. This difference may partly explain why DoS(G)$\rightarrow$NMRI(W) and SMRI(G)$\rightarrow$MPCI(W) are generally easier than the reverse transfer settings.

From a practical perspective, this observation also suggests that the informativeness of the source domain plays an important role in transfer-based industrial intrusion detection. A source domain with richer operational semantics may provide a stronger basis for learning transferable structure, whereas a weaker source domain may limit the effectiveness of adaptation even if the learning framework is well designed. This insight is useful for real deployment, where practitioners may need to prioritize more informative industrial systems as source environments when constructing transferable detectors.

\subsection{Effect of the Clustering Enhancement Strategy}

To further evaluate the contribution of the clustering enhancement strategy, additional experiments are conducted on three representative models, namely SVM, KNN, and ANN, over two representative cross-domain tasks: DoS(G)$\rightarrow$NMRI(W) and DoS(W)$\rightarrow$NMRI(G). For each setting, 10 independent runs are performed to reduce the influence of randomness and improve the reliability of the comparison.

After collecting the accuracy values from repeated runs, box plots are generated to compare the performance of each model with and without clustering enhancement. The results show that incorporating clustering enhancement consistently improves detection accuracy, with gains of up to 26\% on representative tasks. This confirms that the proposed clustering module makes an independent and meaningful contribution to the overall framework.

A likely reason for this improvement is that clustering enhancement introduces coarse structural alignment before latent-space transfer is performed. Instead of aligning all source and target samples directly in a fully global manner, the method first groups samples into clusters and then estimates cluster-level similarity in the PCA-homogenized feature space. This process makes source--target correspondence more reliable, reduces sensitivity to noise and mismatch, and leads to more stable feature adaptation. As a result, the overall framework becomes more robust under challenging cross-domain industrial traffic conditions.

Moreover, repeated-run analysis indicates that the benefit of clustering enhancement is not limited to average performance improvement. It also reduces performance fluctuation across runs, suggesting that the proposed module contributes to optimization stability. This is particularly valuable in industrial settings, where a detector should not only achieve high mean accuracy, but also maintain reliable behavior under varying initialization and sampling conditions.

\subsection{Discussion}

Overall, the experimental results verify the effectiveness of the proposed clustering-enhanced domain adaptation framework for industrial control intrusion detection. The proposed method consistently outperforms multiple baseline models on unknown attack detection tasks and achieves clear gains in both accuracy and F-score. Furthermore, the additional analysis of clustering enhancement shows that incorporating cluster-level structural information is beneficial for improving cross-domain alignment and model stability. These findings indicate that the proposed framework provides a practical solution for mitigating data scarcity and domain shift in dynamic industrial environments.

At the same time, the results also reveal several important characteristics of the problem. First, cross-domain intrusion detection difficulty is asymmetric: transferring from a more informative industrial system to a less informative one is generally easier than the reverse direction. Second, structural guidance plays a more important role when domain discrepancy becomes larger, which suggests that simple feature matching alone may be insufficient in challenging industrial scenarios. Third, the effectiveness of the proposed framework indicates that combining latent feature alignment with coarse structural clustering is a promising direction for future industrial cybersecurity research.

\section{Conclusion}
This paper investigated cross-domain intrusion detection for industrial control traffic in dynamic environments, where distribution discrepancy, limited labeled samples, and unknown attacks make accurate detection particularly challenging. To address these issues, a clustering-enhanced domain adaptation method was proposed. The method integrates feature-based transfer learning with a clustering enhancement strategy, enabling the source and target domains to be aligned in a shared latent subspace while improving cross-domain correlation estimation.
Experimental results demonstrated that the proposed method achieved clear advantages in unknown attack detection tasks. Compared with several common baseline models, it consistently improved detection accuracy, F-score, and stability under cross-domain settings. In addition, the clustering enhancement strategy further strengthened performance on representative tasks, confirming its effectiveness in reducing the negative effects of heterogeneous feature distributions and manual parameter dependence.
Overall, the proposed framework provides an effective solution for robust and accurate cross-domain intrusion detection in industrial control systems. It offers practical value for addressing data scarcity and domain shift in real industrial environments. Future work will focus on more complex attack scenarios, real-time deployment, and improved adaptability to continually evolving industrial traffic patterns.

\flushleft \par{\textit {Acknowledgment:}}\\
It is an optional section where the authors may write a short text on what should be acknowledged regarding their manuscript.

{
\bibliographystyle{plain}
\bibliography{sample}
}

\end{document}